\documentclass[letterpaper]{article}

\usepackage{natbib,alifeconf}  
\usepackage{hyperref}
\usepackage{xcolor}
\hypersetup{colorlinks=true, linkcolor=blue, urlcolor=blue, citecolor=gray}
\title{Ortus: an Emotion--Driven Approach to (artificial) Biological Intelligence}
\author{Andrew W.E. McDonald, Sean Grimes,\and David E. Breen \\
\mbox{}\\
Department of Computer Science \\
Drexel University, Philadelphia, PA 19104 \\
\{awm32,spg63,david\}@drexel.edu} 

\begin{document}
\maketitle

\begin{abstract}
 Ortus is a simple virtual organism that also serves as an initial framework for investigating and developing biologically--based artificial intelligence. Born from a goal to create complex virtual intelligence and an initial attempt to model C. elegans, Ortus implements a number of mechanisms observed in organic nervous systems, and attempts to fill in unknowns based upon plausible biological implementations and psychological observations. Implemented mechanisms include excitatory and inhibitory chemical synapses, bidirectional gap junctions, and Hebbian learning with its Stentian extension. We present an initial experiment that showcases Ortus' fundamental principles; specifically, a cyclic respiratory circuit, and emotionally--driven associative learning with respect to an input stimulus. Finally, we discuss the implications and future directions for Ortus and similar systems.

\end{abstract}

\section{Introduction}

While much work has been done to develop artificial intelligence (AI) systems that borrow principles from organic nervous systems, far less has been done that specifically targets the intersection of biology and artificial intelligence such that adherence to biological principles is of primary concern---rather than to the specific applicability of the technology---with the main goal being a virtual system that exhibits biological intelligence (BI).
As our understanding of organic nervous systems, and access to computation power are ever--growing, widespread interest in systems that do exactly this is greatly increasing. Evidence of this lies in DARPA's recent L2M project, which publicizes their search for machines that learn throughout their lives \citep{darpa}.

Researchers in the realm of computational biology and neuroscience have made progress toward developing systems that model specific organisms or neural circuits, as seen in work done with the nematode Caenorhabditis elegans (C. elegans) \citep{Izquierdo2016}. These systems may require too much focus on organism--specific details in order to achieve proper functionality, shifting focus away from creating more generalized neurologically--inspired intelligent systems.

On the other hand, more traditional (application focused) AI research has started taking more inspiration from human learning, such as the development of an auto--encoder augmented by Hebbian learning that decreases the need for an initial supervised--like learning period \citep{Bowren2016}. Further, \citet{Marblestone2016} discuss ways that artificial neural networks (ANNs) can more closely approximate neural functionality.
In the context of biologically--inspired AI, the frameworks underlying these approaches may be too constraining for full exploration of the potential for the field of biologically--based artificial intelligence. 

Recent work at the intersection of these two areas includes \citet{Sinapayen2016}, which investigates the applicability and biological plausibility of spiking neural networks learning by ``stimulation avoidance''. Perhaps the project most closely aligned to Ortus is a biologically inspired neural network modeled off of a honey bee's visual system, which merges biological mechanisms and neural networks \citep{Roper2017}.

\vspace{.1in}
\textbf{Ortus} is an initial implementation of---and framework for creating---virtual life aimed at approximating the intelligence of living organisms.
Born from the study and analysis of C. elegans' connectome and behavior, it aims to strike a balance between biological abstraction, retention of biological fidelity, and computation scalability in order to approximate biological intelligence and learning as closely as possible.
At its core, Ortus is a network of biologically--inspired, non--spiking neurons, capable of forming excitatory, inhibitory, and electrical synapses.
Similar to the way the structure of C. elegans' 302 neuron nervous system is capable of complex behaviors including toxin avoidance, reflexively withdrawing from a ``tap'', and ``remembering'' the temperature at which it found food \citep{Jarrell2012}, Ortus' ``connectome'' (neural structure) enables its inherent functionality.
Once running, Ortus refines its network---similar to the way organic nervous systems adjust themselves (though far simplified, and not yet as dynamic)---based upon Ortus' intrinsic ``understanding'' that certain things are ``good'' and others are ``bad'', with regard to its own longevity.
This understanding is derived from the structure of the nervous system it generates for itself from a set of input definitions.
We stress that Ortus is \textit{not} intended to model C. elegans, nor any other specific organism. Rather, it is designed to further efforts to develop generalized artificial biological intelligence via generally applicable neurophysiological principles.\footnote{Source code for Ortus is available at \url{https://github.com/OrtusProject/Ortus_ECAL_2017}.} It should be noted that Ortus is a work--in--progress, and its current instability weakens its reliability. This will be addressed in future efforts.

\subsubsection{Contextualizing Emotion.}

\citet{Anderson2014} make the case that relatively simple organisms, such as C. elegans or Drosophila (a genus of flies), have ``emotion primitives'' that contribute to ``central states'' of the nervous system and form the neural basis for what we view to be emotional states.
That is, they argue that there are specific neural circuits that encode a nervous system's state which produce the outcomes that are associated with observable emotions in humans.
Further, they suggest that the (at least) two states of arousal in Drosophila, regulated by dopamine, and the approach--versus--avoidance behavior (Darwinian antithetical pairing) with regard to olfactory stimuli found in C. elegans are examples of ``emotion primitives'' that are factors in emotions seen in different animals \citep{Anderson2014}.

When we discuss emotions in Ortus, our assumption is that in complex organisms, the fundamental building blocks for our complex emotional states are different neural circuits that each contribute to varying degrees at different times.
Our implementation scales this idea back to the point that we have two emotional states, ``fear'' and ``pleasure'', each represented by a single neuron. While simple, both the premise and scalability are supported by \citet{Anderson2014}.

The remaining sections of this paper outline Ortus' design and implementation, describe an initial experiment, discuss the implications of this framework, and analyze its shortcomings.

\section{System Design}

As Ortus aims to be a virtual analogy to intelligent life, we tried to only implement functions that either had a known analogous biological process, or which may have an analogous biological implementation that is unknown, but can be defended with anecdotal evidence. Following each Ortus design element (ODE) below, is its biological rationale (BR).

\begin{figure}
\begin{center}
\includegraphics{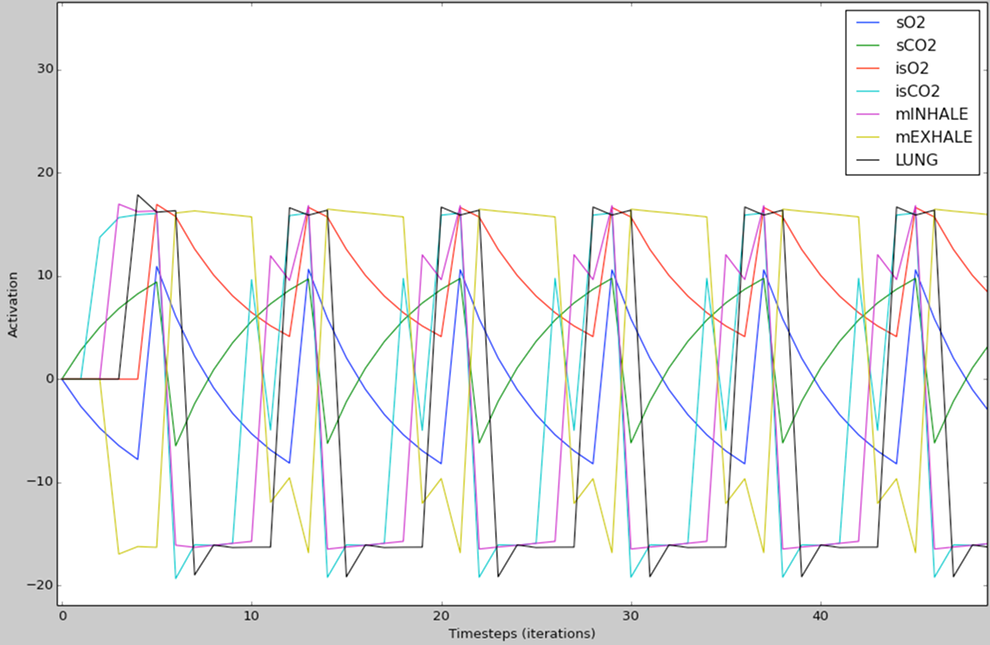}
\caption{Activation cycle of Ortus' respiratory circuit. It can be seen that $sCO_2$ ``naturally'' rises, while $sO_2$ ``naturally'' falls. In response to this, $isCO_2$ (an interneuron linking sensory and motor neurons) increases in activation, which excites the inhalation motor neuron (\texttt{mINHALE}), causing \texttt{LUNG} to activate. This in turn causes $O_2$ to spike, which excites $isO_2$ (an interneuron). The excited $isO_2$ inhibits \texttt{mINHALE} while exciting \texttt{mEXHALE}, thereby allowing $CO_2$ to rise---completing the cycle.}
\label{respiration}
\end{center}
\end{figure}

\textbf{ODE 1:} The underlying source of ``life'' for Ortus is its respiratory circuit, which maintains a balance of $O_2$ and $CO_2$. $O_2$ is consumed, and $CO_2$ is generated, as a product of Ortus being ``alive''. As activation of the $CO_2$ sensory neuron, $sCO_2$, rises and activation of the $O_2$ sensory neuron, $sO_2$, falls, motor neurons responsible for excitation and inhibition of the lung muscle are excited and inhibited (respectively), and the lung increases in activation, supplying $O_2$ and expelling $CO_2$. Fig. \ref{respiration} shows the cyclic nature of the respiratory circuit.

\textbf{BR 1:} Maintaining a given system concentration of $CO_2$ is fundamental to mammalian life, and is very strongly linked to the mammalian fear response (e.g., you get scared if you can't breathe). The ability to regulate one's level of $CO_2$ appears to at least be part of the basis for defining what is a ``good'' and ``bad'' event/stimulus in the mammalian brain.

\textbf{ODE 2:} Currently, Ortus has two ``emotions'', fear and pleasure, represented by emotional interneurons, \texttt{eFEAR} and \texttt{ePLEASURE}, which are both tied into the respiratory circuit. When $CO_2$ rises, \texttt{eFEAR}'s activation rises, and Ortus would then be in a fearful state. As $CO_2$ falls, its contribution to Ortus' fearful state falls. The interaction between $O_2$ and \texttt{ePLEASURE} is the same. In this way, any stimulus presented in combination with either increased $CO_2$ or increased $O_2$ will become known as either a desirable (good) or undesirable (bad) stimulus. Thus, it is via linking new stimuli to known stimuli, that emotions are the driving force behind associative learning in Ortus. In Ortus, the idea of ``emotion'' is simply the rise and fall of activation levels of different neurons or groups of neurons, tied to fundamental behaviors---such as ``breathing''.

\textbf{BR 2:} Clearly, mammals don't normally get scared when they exhale, nor do they feel a measurable increase in pleasure upon inhalation, however the relations do exist on some level. The concept of ``good'' and ``bad'' sensations or emotions only carry meaning to us because of their associations to other neural circuits that are either fundamentally desirable or undesirable from a longevity/survival perspective.
As stated by \citet{Verma2015}, ``emotions, motivations, and reinforcement are a closely related, evolutionarily--conserved phenomena maintaining the integrity of an individual and promoting survival in a natural environment''. Taking this idea into account, along with work by \citet{Gore2015}, which suggests that associative learning is funneled through innate behavior circuits to assign positive or negative emotions to neutral sensory stimuli, it seems that building a virtual organism driven by emotional states is a fairly sound approach. 

While at the human level, the emotional ``part'' of the brain is quite complex, it is not unreasonable to assume that as organismic complexity (and thereby intelligence) decreases, the complexity of emotions decreases. Numerous experiments done on rodents, such as those described by \citet{Weiner2015}, show that the major structures of the brain can be examined by lesioning portions of the brain. Further, it is well known that damage to a specific part of the brain or nervous system can eliminate certain skills, including down to the neuron level when looking at ablation experiments on C. elegans \citep{Fang-Yen2012}. One can infer then, that representing regions of the brain by single neurons would enable a rough approximation of the region's functionality. If one follows this line of thought, the possibility emerges that organisms like C. elegans may, in fact, be driven by ``emotions'' as well. For example, C. elegans is capable of toxin avoidance, a tap--withdrawal response, as well as learning that it found food at a certain temperature \citep{Wicks1996,Xu2012}. One must ask how this can be. There is nothing external that assists it in differentiating good from bad, yet it wants to avoid certain things, while it is attracted to others. In Ortus, we make the assumption that these behaviors are a result of a \textit{very} simple emotional subsystem that forms the basis for C. elegans' behavior.

\textbf{ODE 3:} Ortus employs four different classes of interneurons. First are ``Sensory Extension Interneurons'' (SEIs).\footnote{The two interneurons discussed in Fig. \ref{respiration} are SEIs.} These take input directly from sensory neurons, and pass the input along to the second class of interneurons, ``Sensory Consolidatory Interneurons'' (SCIs).
SCIs may have between 1 and the number of sensory inputs as chemical synapses (CSs), with incoming synaptic weights equal to \textit{1 / (\# sensory inputs)}.
The idea behind SCIs is to enable different types of sensors to combine their input, and trigger emotions, effectively as a ``new'' sensory input, thereby forming associations between two stimuli. We defer the descriptions of the last two interneurons to \textbf{ODE 5}.

\textbf{BR 3:} Admittedly, SEIs may not be necessary. We included these to easily give sensory neurons a functionality that was separate from interneurons if the need arises. With regard to SCIs, it is suggested by \citet{Xie2016} that neurons in the brain are organized according to the idea that if there are $N$ neurons, then the brain has the ability to represent all $2^N-1$ possible combinations.
Clearly, for anything but the most simple organisms, having one neuron that collects the input for each of the $2^N-1$ possibilities is unrealistic (as the authors note).
However, the authors suggest there may be additional combining of neuron inputs to decrease the computational and spacial complexity (in organic brains, that is).
It also possible that mammalian brains are not quite as connected as they seem to think they are, and a great deal of what the brain does amounts to interpolation.
Regardless, we employed the idea of $2^N-1$ SCIs because losing sensory resolution with such few neurons (that only have one neuron per sensory input) doesn't make sense.
This will have to be reassessed as the system develops.
There is, however, a far stronger argument for the strength of the SCI inputs: it seems that synaptic strength scales inversely with the number of connections, ``K as ~ 1 /$\sqrt{K}$'' \citep{Barral2016}. This makes intuitive sense; if neuron \texttt{A} synapses onto \texttt{B} with a weight of $1$, and \texttt{C, D, E} all synapse onto \texttt{F}, the weight of each of the latter three connections must be $1 \over 3$ in order to maintain equivalence between \texttt{B} and \texttt{F}. In addition, there is evidence that neighboring neurons in the same ``layer'' are connected in C. elegans \citep{Azulay2016}, which suggests that a certain amount of information consolidation may occur.

\textbf{ODE 4:} Ortus currently implements both Hebbian learning and the Stentian extension to Hebbian learning. Specifically, for each chemical synapse, on every timestep, if activity is sufficiently synchronous or sufficiently asynchronous, the synapse strengthens or weakens (respectively) according to its \textit{mutability index} (\textit{MI}). The \textit{MI} determines a synapse's potential to be modified. Currently, \textit{MI}s are static, though in the future we plan to vary them with synaptic age, among other things.

\textbf{BR 4:} Hebbian and Stentian learning, relating to the correlation (or lack thereof) between presynaptic and postsynaptic pairs is a proven learning paradigm in neuroscience, as described by \citet{Kutsarova2016}.

\begin{figure}[t]
\begin{center}
\includegraphics{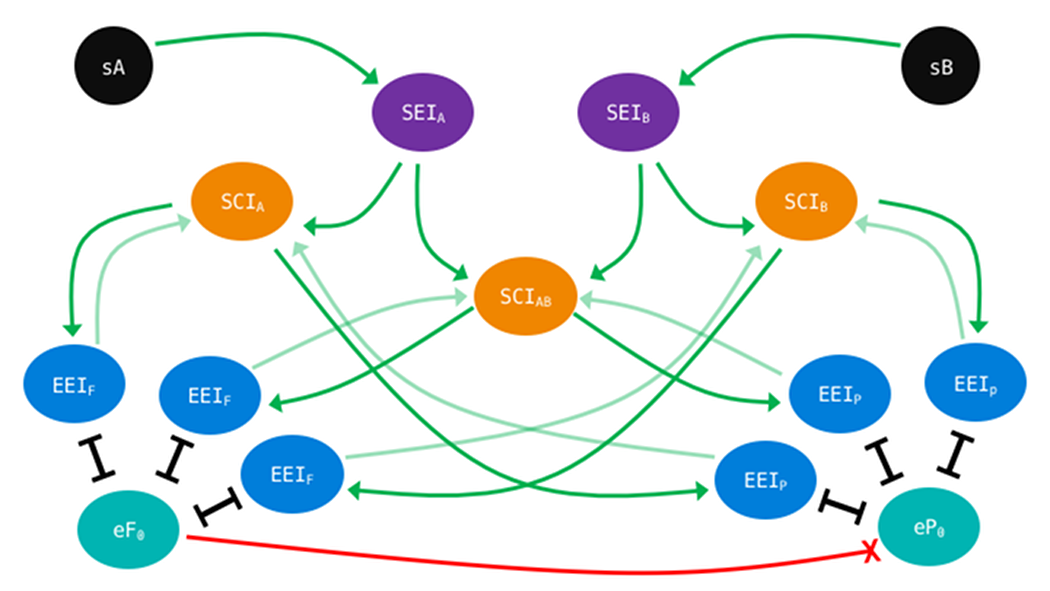}
\caption{Diagram of Ortus' emotional integration. Each color (other than black, which are sensory neurons) represents a specific layer of interneurons. SEIs are purple, SCIs are orange, EEIs are blue, and primary emotions ($eF_0$ and $eP_0$) are teal. Dark green arrows represent excitatory chemical synapses, while light green arrows represent weak chemical synapses. Black bars with flat caps on each end depict bidirectional gap junctions, and the red connection with an ``X'' at the end, symbolizes an inhibitory (``opposing'') relationship, in which not only does $eF_0$ inhibit $eP_0$, but the group of EEI$_F$s inhibit the EEI$_P$s connected to their respective SCIs. This is the basic structure that Ortus employs. Currently there is no restriction regarding which interneurons can connect to motor neurons.}
\label{diagram}
\end{center}
\end{figure}

\textbf{ODE 5:} The third class of interneurons is comprised of ``Emotion Extension Interneurons'' (EEIs), which are essentially auxiliary emotion neurons.
EEIs are specific to each emotion, and receive input from SCIs via chemical synapses, and pass that input on to their respective primary emotion neurons via gap junctions (GJs).
Emotional learning is achieved by strengthening synapses (via Hebbian learning) of interneurons that form synapses between SCIs (consolidated sensory information), and EEIs.
The GJ connection enables a given emotional state, incurred by any stimulus, to permeate through Ortus---resulting in an ``emotional state'' of the system (gap junctions in Ortus have no activation threshold).
The effect of this is that, if a certain sensory input causes a given emotion to increase in activation, the introduction of another sensory input will cause the synapses at the junction of the newly introduced sensory input and elevated emotion to strengthen. 
Further, as each SCI synapses onto one EEI per emotional state, each EEI has a fairly weak excitatory chemical synapse to their parent SCI.
The idea behind the synapse from the EEI to the SCI is to allow an emotional state to ``trigger'', or cause Ortus to ``remember'' a stimulus that previously invoked (and thus become associated with) that emotional state.
In this case, ``remembering'' is defined as measurably increased activation in the SCI that the EEI synapses onto.
In practice, we have observed a very slight difference in Ortus as a result of these connections; currently they exist more as a concept that does not break the system. A diagram of this layout may be seen in Fig. \ref{diagram}.

\textbf{BR 5:} Specifically with regard to the mammalian dopamine circuit, there exists a small locus of dopaminergic cells that receive inputs from diverse sources and project to diverse parts of the brain, as discussed by \citet{Beier2015}. From a psychological perspective, it is clear that when one enters an emotional state, the emotion is globally encompassing rather than localized. It is also clear that a stimulus or activity is likely to be ``colored'' by the emotional state one was in when the stimulus occurred. The approach we took with Ortus is a simplification of the dopamine circuit, and a generalization across emotional states; it will likely need refinement.

\textbf{ODE 6:} Certain emotional states preside over (dominate) others. In Ortus, fear dominates pleasure.

\textbf{BR 6:} The idea of a hierarchical system, where certain emotional states dominate others, is supported by the inhibition of fear in mice, in favor of searching for food when hungry (blood glucose levels falling causing the release of hormones), but greater concern for safety when not hungry \citep{Verma2015}. Further, \citet{Leknes2008} discuss the ``Motivation--Decision Model'', which suggests that anything that is more important for survival than pain should inhibit the feeling of pain.

\textbf{ODE 7:} Ortus builds its nervous system by reading in a \textit{.ort} file, using our ``\textit{Ortus Development Rules}'' language. First, ``elements'' (neurons and muscles) must be specified, along with attributes such as the type of element, its affect, and activation threshold. Once all elements have been defined, relationships between elements may be specified. Currently, there are four relationships:
\begin{itemize}
    \itemsep0em
    \item \texttt{[+-]A} \textbf{causes} \texttt{[+-]B}
        \subitem Where the ``+''/``-'' indicates an increase/decrease in the presynaptic neuron causes an increase/decrease in the postsynaptic element. This translates into a chemical synapse.
    \item \texttt{A} \textbf{correlated} \texttt{B}
        \subitem Where A and B are correlated. This translates into a bidirectional gap junction.
    \item \texttt{A} \textbf{opposes} \texttt{B}
        \subitem Where A and B inhibit each other.
     \item \texttt{A} \textbf{dominates} \texttt{B}
         \subitem Where A inhibits B.
\end{itemize}

Relationship--level attributes may be specified, such as: ``mutability'', ``weight'', and ``polarity''. At this time, most relationship attributes need not be specified as Ortus uses default values for mutability and weight, while polarity is determined from the relationships defined. Once Ortus reads these instructions in, it creates interneurons and connections as described by the \textbf{ODE}s above to satisfy all constraints imposed, and to allow associative learning between all emotional states and sensory stimuli (both individual and combined) as described in \textbf{ODE 5}.

\textbf{BR 7:} Our goal in creating the \textit{Ortus Development Rules} was to approximate, in a greatly simplified manner, the way a brain grows itself, based upon genetic instructions that cause circuits to form in certain ways via gene expression \citep{Weiner2015}. Ortus' neural structure is essentially wired to force it to behave in certain ways, akin to the innate behaviors observed in mammals. Further, \citet{Schroter2017} provide evidence for the existence of organizational ``motifs'' found in C. elegans that may underly more complex networks in larger brains; this also lends credence to Ortus' rule--based development approach.

\subsection{Implementation Details} Ortus is written in C++ and OpenCL. Initial development of its neural structure and sensory stimulation are carried out through C++, while all signal transmissions between neurons and learning is done in the OpenCL kernel. Each iteration of the OpenCL kernel constitutes one timestep, during which each neuron sums incoming (positive or negative) ``activation'' from presynaptic cells via chemical synapses and gap junctions. We use the term ``activation'' in place of ``potential'' or ``voltage'' to make it clear that we aren't dealing with the transfer of electrical current, but have abstracted that away to simply ``activation''. The chemical synapse and gap junction activation transfer equations below were adapted and simplified from those described by \cite{Wicks1996} to ignore physical properties of neurons (such as neuron length), decrease computational complexity, and reduce potential sources of hard--to--trace error.

Ortus' neurons are based upon C. elegans' neurons, and are non--spiking, but do not transmit any ``activation'' below a given threshold (for chemical synapses, gap junctions have no threshold), as described by \citet{Graubard1014}. The equations governing activation transmission between Ortus' neurons are as follows (unless otherwise specified, $A$ refers to the postsynaptic neuron, $i$):

\begin{equation}
    A[m+1] = (A[m] - A_{decay}) + A_{GJ_{in}} + A_{CS_{in}},
\end{equation}

where $A[m]$ and $A[m+1]$ are the activations of neuron $i$ for the current and next timestep, $A_{decay}$ is the amount of activation lost by neuron $i$ during the current timestep, and $A_{GJ_{in}}$ and $A_{CS_{in}}$ are the ``incoming'' activations to neuron $i$ from gap junctions and chemical synapses (respectively).

\begin{equation}
    A_{GJ_{in}} =  GJ_{pre,post} {(A_{pre}[m] - A[m]) \over 2},
\end{equation}

where $GJ_{pre,post}$ is the weight of the chemical synapse from the $pre$--synaptic to $post$--synaptic neuron, $A_{pre}[m]$ is the presynaptic neuron's activation. We divide the activation difference by 2 because only half of the difference should move into the $i^{th}$ neuron; $pre$ will lose that half to $i$, and will keep the other half. This is analogous to pouring water into one of two water tanks joined at the bottom.

\begin{equation}
    A_{CS_{in}} =  CS_{pre,post} S_g  (A_{rev} - A[m]),
\end{equation}

where $CS_{pre,post}$ is the weight of the chemical synapse from the $pre$--synaptic to $post$--synaptic neuron, $S_g$ is the synaptic conductance of synapse $CS_{pre,post}$, and $A_{rev}$ is the reversal activation for the synapse (will be negative if an inhibitory synapse, and positive if excitatory).

\begin{equation}
    \label{eq:conductance}
    S_g = {1 \over { 1 + e^{(-5 { A_{pre}[m] \over A_{range} } )} }},
\end{equation}

where $A_{range}$ is the activation range of the presynaptic neuron (which is the difference between the excitatory and inhibitory reversal potentials). Eq. \ref{eq:conductance} uses -5 to vary the conductance from a little less than 10\% to a little more than 90\%.
It should be noted that generally we would subtract the equilibrium activation, $A_{eq}$, from $A_{pre}[m]$ prior to dividing by $A_{range}$, but this is not necessary because $A_{eq} = 0$.
We use the presynaptic neuron's activation to determine the synaptic conductance because our neurons are non--spiking, and as such, activate their synapses with magnitude  proportionate to the neuron's activation level \citep{Wicks1996}.

\begin{equation}
    A_{decay} = (C_D A[m]) - A_{GJ_{out}}[m],
\end{equation}

where $C_D$ is a constant decay percentage (set to 20\%), and $A_{GJ_{out}}[m]$ is the amount of activation neuron $i$ will lose from outgoing gap junctions.

\begin{equation}
    A_{GJ_{out}} = \sum_{j=1}^{N}{(GJ_{out_j} {(A[m] - A_j[m]) \over 2 })},
\end{equation}

where $GJ_{out_j}$ is the weight of neuron $i$'s $j^{th}$ outgoing gap junction, $N$ is the number of neurons,  and $A_j[m]$ is the $j^{th}$ neuron's activation. Note that to compute the activation that neuron $i$ will lose due to its outgoing gap junctions, we treat it as the presynaptic neuron. 

\subsection{Learning}

In order to implement biologically plausible Hebbian and Stentian learning, synchronous activity had to be locally determined on a per--neuron basis. Each postsynaptic neuron computes a cross--correlation between its previous four activations and all other neurons from the current time to four timesteps back, as seen in Eq. \ref{eq:xcorr}.

\begin{equation}
    \label{eq:xcorr}
    XCorr[t] =  xcorr(H_{post}[0..3], H_{j}[t..t+4]),
\end{equation}

where $XCorr[t]$ is the $t^{th}$ cross--correlation computation between $post$ and presynaptic neuron $j$, $H$ is the activation history indexed by most recent activation at $H[0]$, and $xcorr$ is the cross--correlation computation.

This allows a postsynaptic neuron \texttt{A} to determine how every other neuron is moving with respect to itself at the current time ($t=0$), and how their activity may have \textit{influenced} it, by assessing the level of correlation between progressively earlier historical activations of the other neurons relative to itself. A high correlation at $XCorr[3]$, for example, would indicate that the $post$ neuron acted similarly to neuron $j$, three timesteps after neuron $j$ acted that way.

Neurons also compute the slopes of the activations of all neurons, as shown in Eq. \ref{eq:slope}, which allows detection of a constant or rapidly changing neuron's activation.

\begin{equation}
    \label{eq:slope}
    Slope_j[t] =  slope_u(H_{j}[t..t+u]),
\end{equation}

where $Slope_j[t]$ is the $t^{th}$ slope computation for neuron $j$, $u$ is the number of timesteps to use when computing the slope (set to 2), and $slope$ computes the slope.

Combining the $XCorr$ and $Slope$ computation results, assuming both neurons involved have activations above the activation threshold, produces the following results (note that the maximum $XCorr$ value is 1).

\begin{itemize}
    \itemsep0em
    \item Rapid strengthening occurrs if there is a highly synchronous pair of nearly constant signals:
        \subitem $\sum_{i=1}^{4}{ XCorr[i]} \ge 3.92$, and
        \subitem $\sum_{i=1}^{4}{ Slope[i]} \le .02$,
    \item Synaptic weakening occurrs slowly if:
        \subitem $\sum_{i=1}^{4}{ XCorr[i]} < .05$,
    \item Synaptic strengthening occurrs slowly if:
        \subitem $\sum_{i=1}^{4}{ XCorr[i]} > 3.5$,
\end{itemize}

\section{Experimental Design and Results}

\subsubsection{Experiment.} To test our initial implementation of Ortus, we implemented the configuration described above, with the addition of an $H_2O$ sensory neuron, $sH_2O$, and corresponding interneurons..
We then ``exposed'' Ortus to water via $sH_2O$ stimulation four times in bursts spaced 100 timesteps apart, in order to allow injected activation to naturally decay.
During each burst of $H_2O$ exposure, we prevented Ortus from exhaling $CO_2$ or inhaling $O_2$, thereby inducing an enhanced state of fear.
After repeating the conditioning exposure four times, we allowed Ortus 200 timesteps to let any injected activation to decay.
Finally, we exposed Ortus to water, without inducing a fearful state, to determine if it had learned to be fearful of water.

\subsubsection{Results.} 
\begin{figure*}
\begin{center}
\includegraphics{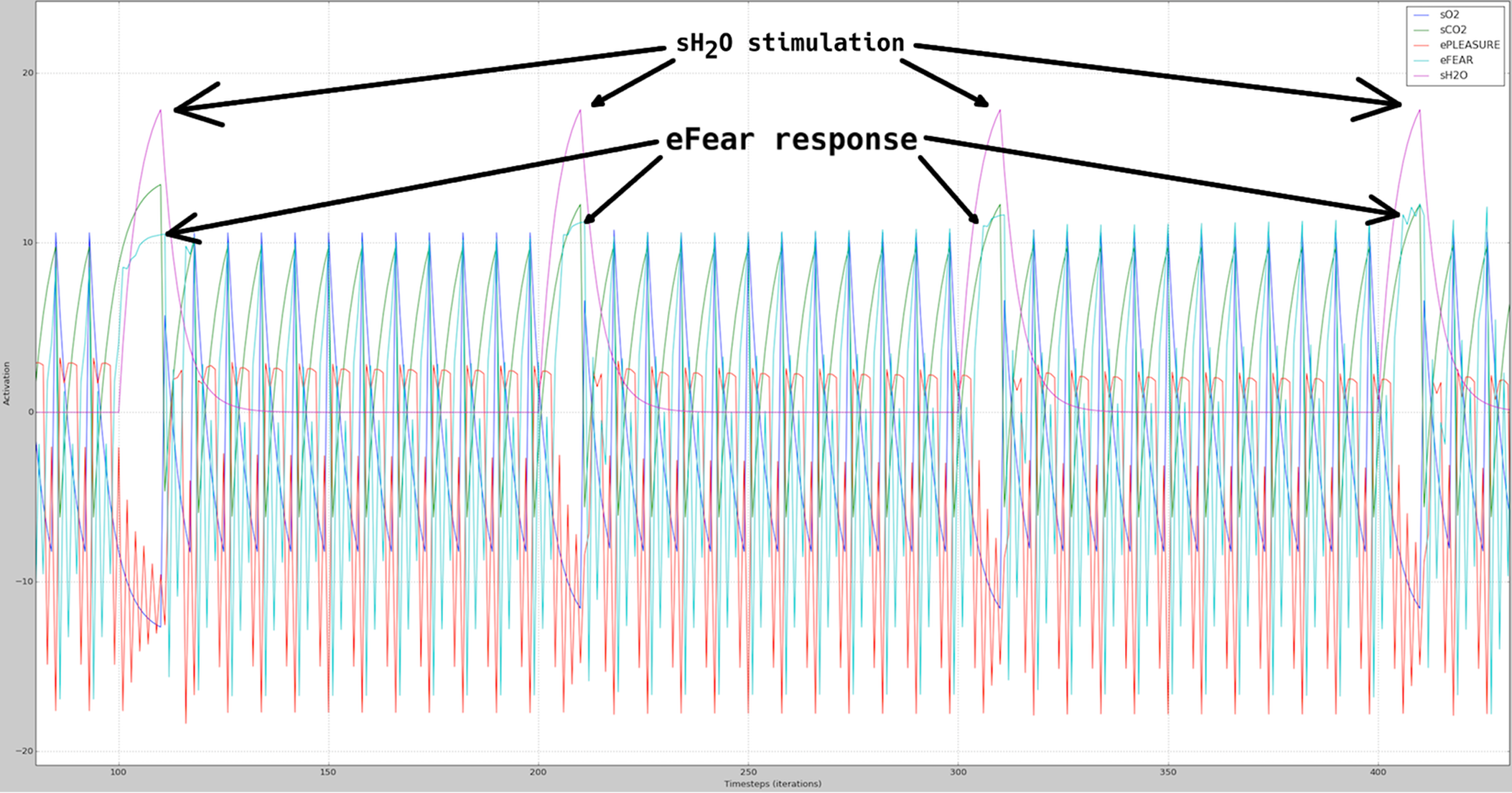}
\caption{Ortus' fear conditioning with $CO_2$ exhalation prevention and $O_2$ deprivation. Each time $H_2O$ is introduced under these conditions, Ortus' fear response grows, as evidenced by a strengthening response of \texttt{eFEAR} during each of the four rounds of conditioning.}
\label{conditioning}
\end{center}
\end{figure*}

\begin{figure}
\begin{center}
\includegraphics{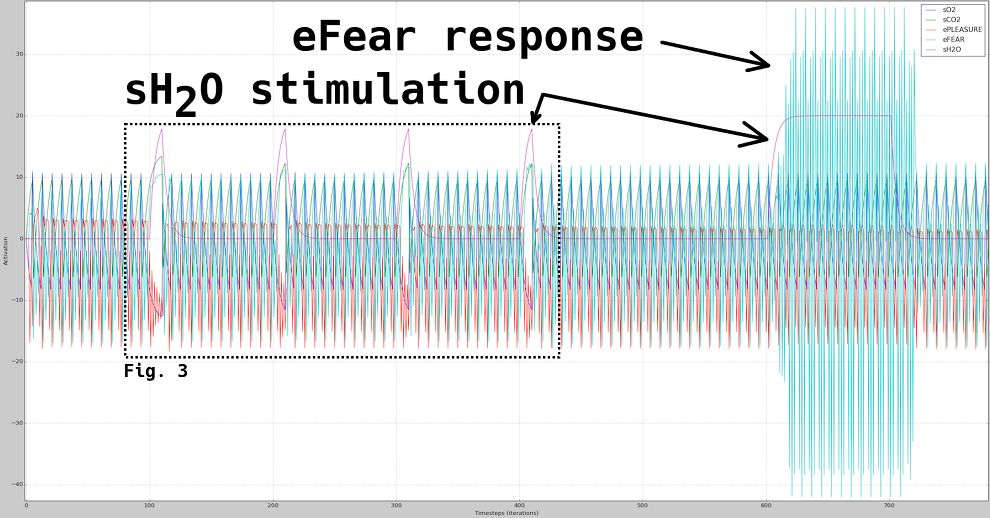}
\caption{The final result of conditioning Ortus to fear $H_2O$. After the four rounds of conditioning, simply exposing Ortus to $H_2O$ is enough to induce a state of fear.}
\label{conditioned}
\end{center}
\end{figure}

\begin{figure}
\begin{center}
\includegraphics{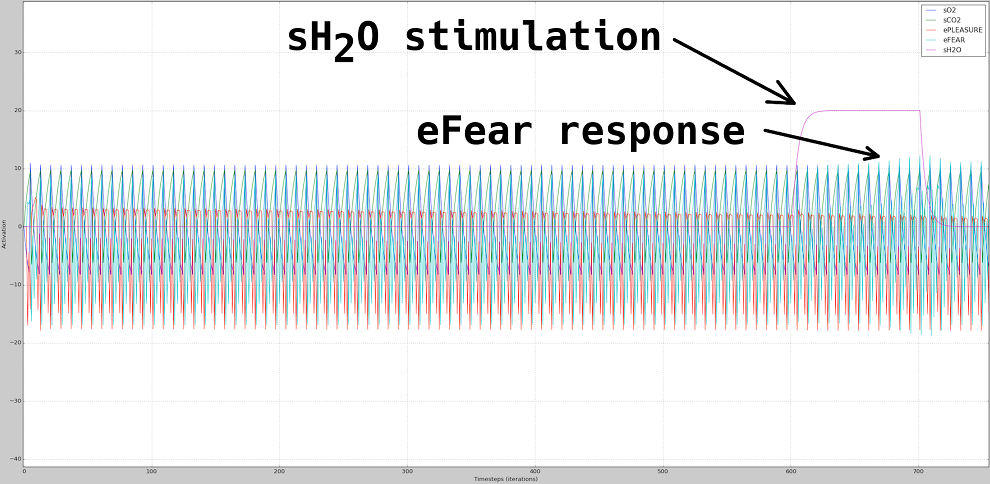}
\caption{In the absence of fear conditioning, $H_2O$ introduction causes a very subtle, slowly growing association with fear. This is likely due to instability in Ortus, and is an indication Ortus' learning needs more nuance.}
\label{no_conditioning}
\end{center}
\end{figure}

Since $sCO_2$ activation causes fear, and $H_2O$ was presented ($sH_2O$ was stimulated) during this time, we expected Ortus to become progressively more and more ``scared'' of water each time it was presented while it was in a state of elevated fear.
Specifically, we wanted to use classical conditioning to teach Ortus to fear water. 
As can be seen in Fig. \ref{conditioning}, during each exposure to $H_2O$, activation of \texttt{eFEAR} increased when compared to the previous exposure.
In Fig. \ref{conditioning} the sharply increasing purple trace is the $H_2O$ sensor, the dark green trace just below it is the $CO_2$ sensor, and the light blue trace below that is the fear response.
After the four rounds of conditioning, we expected to observe an increase in activation in \texttt{eFEAR} when we presented Ortus with $H_2O$, in the absence of exhalation and inhalation restriction. Fig. \ref{conditioned} confirms this increase in fear activation as a result of $sH_2O$ stimulation alone. Compared to simply exposing Ortus to $sH_2O$ stimulation without prior fear conditioning, in which a very slight raise was observed---as may be seen in Fig. \ref{no_conditioning}, it is clear that we were able to successfully use classical conditioning to instill a fear of water in Ortus that was previously almost non--existent.

\setlength{\belowcaptionskip}{-.25in}

\section{Discussion}

Our initial experiments with Ortus show that it has potential as a biologically--based framework for developing virtual life that exhibits complex intelligence. That said, there are a number of shortcomings that must be addressed in order for Ortus to become much more complex.

First, there can be instability in the system. Further development, analysis, and experimentation of the emotional circuits are necessary. The current approach is straightforward, and seems to play a role in the observable instability. A number of additional factors could play a role in its instability as well; at this point, it is quite easy to throw Ortus out of balance. Small changes to the weights can have significant implications for the functionality of the system because an initial small change can multiply causing the system to become unstable. But it should also be noted that only a few hours of experimentation were needed to find the system parameter values that produced our results. \citet{Johansen2014} suggests that in mice, while synchronous activity (correlation) is necessary for synaptic plasticity, it is not sufficient; neuromodulatory signaling is also required. This could be a way to control some of the unstable behavior seen as a result of the implemented purely correlation--based learning rules. In general, implementing more biologically--based constraints should help solve this, though refinement of the neural structuring rules is also likely necessary.

In fact, the neural structure needs refinement for combinatoric reasons anyway; in order to implement a sensor array (e.g., any sort of topographically ordered sensory inputs, like vision), using a ``fully--wired'' approach is completely unfeasible. Even 32 sensors would require over 4 billion neurons, if each possible combination of sensory inputs mapped to a different neuron. \citet{Schroter2017} suggests that C. elegans' neurons may have multiplexed functions, meaning that one neuron may contribute to more than one behavior. This is another possible way to decrease the number of necessary neurons in Ortus.

Lastly, we would like to make Ortus' learning far more nuanced, and capable of picking up on more subtle changes than it currently can; though working on resolving the aforementioned issues should help the learning aspect as well. Additionally, \citet{Kutsarova2016} indicate that, relative to synaptic strengthening and weakening, axonal branch tips emerge to form new synapses, and that synchronous activity stabilizes synapses and prolongs axonal branches. In addition to refining Ortus' method for altering its synaptic weights, this is one more way for Ortus to modify its approach to learning. It also suggests there is biological precedent to implement a ``floor'' so that as Ortus learns new things, they can't be unlearned beyond a certain point.

Shortcomings aside, the results of our classical conditioning experiment with Ortus show that we successfully created a biologically--based virtual system that has a ``self--sustaining'' respiratory circuit, and is capable of Hebbian and Stentian learning.
We have made the case that Ortus is emotion--driven, and have shown that we can both induce fear in it by interrupting its respiration cycle, and teach it to fear things it hadn't previously feared.
To suggest Ortus was virtually ``alive'' at any point during our experimentation would be a stretch, however, once Ortus and similar systems increase in complexity, we must pose the question: when can a virtual organism ``feel'', if ever?

\section{Conclusion}
We have presented Ortus, an initial approach to a framework that exhibits basic emotionally--based learning (fear conditioning), self--sustaining inherent behavior (cyclic breathing via a respiratory circuit), and an approach to sensory processing that enables inter-sensory associative learning, with or without an emotional component. While the implementation of Ortus presented is in its initial stages, the design principles are solidly based in biology, and with further development and refinement, the potential exists for Ortus' development of (artificial) biological intelligence to quickly increase. 

\section{Acknowledgements}

We thank Denisa Qori for her helpful thoughts, and Rachel Greenstadt for her prior guidance.
This work was supported by the NSF Graduate Research Fellowship Program. 

\footnotesize
\bibliographystyle{apalike}
\bibliography{ortus_paper} 


\end{document}